\documentclass[10pt]{cai}
\usepackage{acronym}
\newacro{DNN}{Deep Neural Network}
\newacro{CNN}{Convolution Neural Network}
\newacro{CAM}{Class Activation Mapping}
\newacro{SODEx}{Surrogate Object Detection Explainer}
\newacro{D-RISE}{Detector Randomized Input Sampling for Explanation}

\begin{document}
\def\conferenceyear{2024}
\volumeheader{37}{0}
\begin{center}

\title{Efficient and Concise Explanations for Object Detection with Gaussian-Class Activation Mapping Explainer}
\maketitle

\thispagestyle{empty}


\begin{tabular}{cc}

Khanh Nguyen\upstairs{\affilone}, Hung Nguyen\upstairs{\affilone,\affiltwo,*},\\
Khang Nguyen\upstairs{\affilone}, 
Binh Truong\upstairs{\affilone},
Tuong Phan\upstairs{\affilone,\affilthree},
Hung Cao\upstairs{\affiltwo}
\\[0.25ex]
{\small \upstairs{\affilone}Quy Nhon AI, FPT Software, Vietnam} \\
{\small \upstairs{\affiltwo}Analytics Everywhere Lab, University of New Brunswick, Canada} \\
{\small \upstairs{\affilthree}University of Waterloo, Canada} \\
\end{tabular}
  
\emails{
  \upstairs{*}hung.ntt@unb.ca 
}
\vspace*{0.2in}
\end{center}

\begin{abstract}
To address the challenges of providing quick and plausible explanations in Explainable AI (XAI) for object detection models, we introduce the Gaussian Class Activation Mapping Explainer (G-CAME). 
Our method efficiently generates concise saliency maps by utilizing activation maps from selected layers and applying a Gaussian kernel to emphasize critical image regions for the predicted object. 
Compared with other Region-based approaches, G-CAME significantly reduces explanation time to 0.5 seconds without compromising the quality. 
Our evaluation of G-CAME, using Faster-RCNN and YOLOX on the MS-COCO 2017 dataset, demonstrates its ability to offer highly plausible and faithful explanations, especially in reducing the bias on tiny object detection.
\end{abstract}

\copyrightnotice

\section{Introduction}\label{sec:introduction}
In object detection, \acp{DNN}~\cite{girshick2014rich} have significantly improved with the adoption of \acp{CNN}.
However, the deeper the network is, the more difficult it is to understand, debug, or improve, which potentially poses a serious problem in critical areas~\cite{nguyen2023towards}.
To help humans gain a thorough understanding of the model's decisions, several Explainable Artificial Intelligence (XAI) methods using saliency maps to highlight the important regions of input images have been introduced.

A common and simple way to explain the object detector is to disregard the model's architecture and only consider the input and output. 
This approach aims to determine the importance of each region in the input image based on the change in the model's output. 
For example, \ac{D-RISE}~\cite{petsiuk2021black} estimates each region's effect on the input image by creating thousands of perturbed images, and subsequently feeding them into the model to predict and get the score for each perturbed mask.
Another method is \ac{SODEx}~\cite{sejr2021surrogate}, an upgrade of Local Interpretable Model-Agnostic Explanations (LIME)~\cite{ribeiro2016should}, which also employs the same technique as D-RISE to explain object detectors.
Although the results of both SODEx and D-RISE are compelling, the generation of a large number of perturbations slows the explanation generation considerably.

Other approaches, such as \ac{CAM}~\cite{zhou2016learning} and GradCAM~\cite{selvaraju2017grad}, use the activation maps of a specific layer in the model's architecture as the main component to form the explanation.
These methods are faster than the mentioned region-based but still have some meaningless information since the feature maps are not related to the target object~\cite{zhang2021group}. 
Such methods can give a satisfactory result for the classification task. Still, they cannot be applied directly to the object detection task
because these methods highlight all regions having the same target class and fail to focus on one specific region.

In this paper, we propose \textit{Gaussian Class Activation Mapping Explainer} (G-CAME), which can explain the classification and localization of the target objects. 
Our method extends the applicability of CAM-based XAI to object detectors. 
By adding the Gaussian kernel as the weight for each pixel in the feature map, G-CAME's final saliency map can explain each specific object. 
Our contributions can be summarized as follows:
\begin{enumerate}
    \item We propose the first CAM-based method tailored for object detection, G-CAME, which can explain object detectors as a saliency map for a specific target object.
    G-CAME can explain in a reasonably short time, which overcomes the existing methods' time constraints like D-RISE~\cite{petsiuk2021black} and SODEx~\cite{sejr2021surrogate}.
    \item We qualitatively and quantitatively evaluate our method with D-RISE on two main types of object detectors, namely YOLOX~\cite{ge2021yolox} (one-stage detector) and Faster-RCNN~\cite{ren2015faster} (two-stage detector), and prove that our method can give a less noise, more accurate saliency map in a shorter time than D-RISE.
\end{enumerate}
Our code is available at \url{https://github.com/khanhnguyenuet/GCAME}.

\section{Explainable AI in Object Detection}\label{sec:related-work}
Object detection, a field in computer vision (CV), involves models that are broadly classified into two categories: one-stage and two-stage models. 
One-stage models, such as the YOLO series~\cite{redmon2016you}, SSD~\cite{liu2016ssd}, and RetinaNet~\cite{lin2017focal}, detect objects directly over a dense sampling of locations. In contrast, two-stage models like the R-CNN family~\cite{girshick2014rich}, FPN~\cite{lin2017feature}, and R-FCN~\cite{dai2016r}, involve a two-phase process. Initially, these models select Regions of Interest (ROI) from the feature extraction stage, followed by classification based on each proposed ROI.

While several XAI methods have been applied to analyze deep \ac{CNN} models in classification tasks, their applicability in object detection is comparatively limited due to constraints in flexibility, suitability, and computational efficiency~\cite{8689279}.

This section discusses two XAI types: Region-based saliency methods and CAM-based saliency methods. These methods are evaluated for their applicability in both classification and object detection tasks. 
A significant gap in current XAI methods, particularly in object detection, is identified, laying the groundwork for the introduction of our method.

\subsection{Region-based saliency methods}
Region-based saliency methods use masks to isolate specific regions of an input image, assessing their impact on the output by processing the masked input through the model and quantifying each region's influence. 
In classification, LIME~\cite{ribeiro2016should} and its extension, RISE~\cite{petsiuk2018rise}, are notable examples, where the latter employs thousands of masks to generate a composite saliency map. 
Recent advancements have adapted these methods for object detection.
SODEx~\cite{sejr2021surrogate} applies LIME to explain object detectors, modifying the metric to focus on target bounding boxes. 
D-RISE~\cite{petsiuk2021black} refines this by altering the computation of weighted scores for each random mask, specifically for object detection.
D-CLOSE~\cite{truong2023towards} further utilizes multiple levels of
segmentation on the image and combines them to deliver more concise and consistent explanations.
Region-based methods offer an intuitive approach as they do not necessitate the end-users in-depth understanding of the model's architecture.

However, a notable challenge is the sensitivity of these explanations to changes in hyper-parameters, resulting in multiple potential explanations for a single object. 
Consequently, to achieve a clear and satisfactory explanation, careful fine-tuning hyper-parameters is essential. 
Additionally, a significant drawback of region-based methods is the considerable amount of time required to generate an explanation.

\subsection{CAM-based methods}
Conversely, CAM-based XAI requires a thorough understanding of the model's architecture. Techniques such as CAM~\cite{zhou2016learning} and its successors, GradCAM~\cite{selvaraju2017grad}, GradCAM++~\cite{chattopadhay2018grad}, and XGradCAM~\cite{fu2020axiom}, are noteworthy for producing detailed saliency maps. 
These methods utilize partial derivatives of feature maps in selected layers relative to the target class score. 
While CAM-based methods are generally more efficient than Region-based methods~\cite{nguyen2021evaluation}, their reliance on feature maps can result in less meaningful saliency maps.
Additionally, these methods have primarily been developed for classification tasks, with no existing adaptations for object detection.

In light of these limitations, we introduce G-CAME, a novel CAM-based XAI method tailored for object detection. 
G-CAME is the first of its kind to offer stable and rapid explanations for both one-stage and two-stage object detection models, addressing the shortcomings of existing approaches.

\section{Proposed method}\label{sec:proposed-methods}
\begin{figure*}[h!]
    \centering
    \includegraphics[width=.9\linewidth]{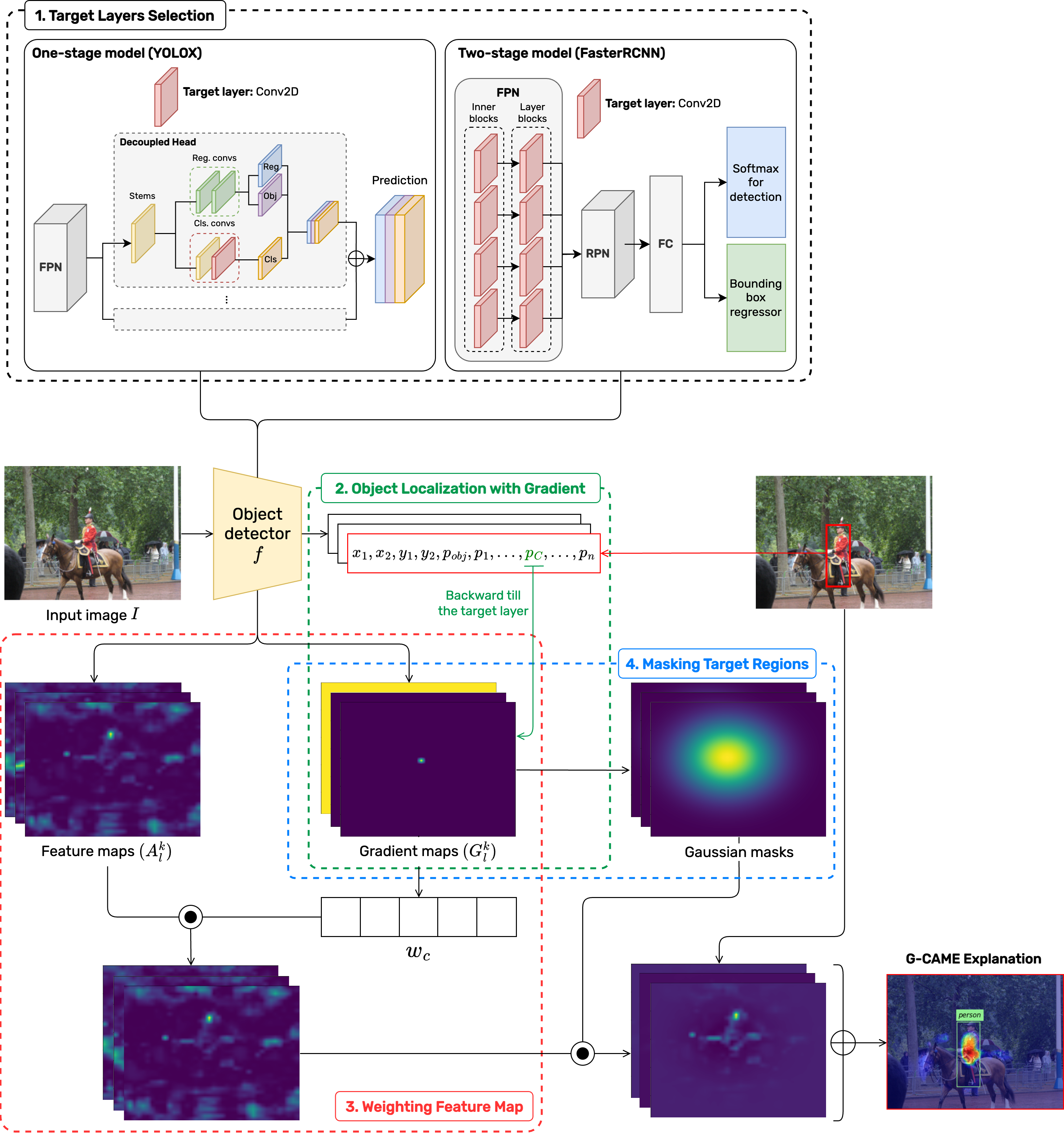}
    \caption{Overview of G-CAME method. We use the gradient-based technique to get the target object's location and weight for each feature map. 
    We multiply element-wise with Gaussian kernel for each weighted feature map to remove unrelated regions. 
    After applying the Gaussian kernel, the output saliency map is created by a linear combination of all weighted feature maps.}
    \label{fig:method_overview}
\end{figure*}

For a given image $I$ with size $h$ by $w$, an object detector $f$ and the prediction $d$ includes the bounding box and predicted class. 
We aim to provide a saliency map $S$ to explain why the model has that prediction.
The saliency map $S$ has the same size as the input $I$.
Each value $S_{(i, j)}$ shows the importance of each pixel $(i, j)$ in $I$, respectively, influencing $f$ to give prediction $d$. 
We propose a new method that helps to produce that saliency map in a white-box manner.
Our method is inspired by GradCAM~\cite{selvaraju2017grad}, which uses the class activation mapping technique to generate the explanation for the model's prediction.
The main idea of our method is to use normal distribution combined with the CAM-based method to measure how one region in the input image affects the predicted output. 
Fig.~\ref{fig:method_overview} shows an overview of our method.

Due to their output difference, we cannot directly apply XAI methods for the classification model to the object detection model.
In the classification task, the model only gives one prediction that shows the image's label. 
However, in the object detection task, the model gives multiple boxes with corresponding labels and the probabilities of objects.
Most object detectors, such as YOLO~\cite{redmon2016you} and R-CNN~\cite{girshick2014rich}, usually produce $N$ predicted bounding boxes in the format:
\begin{equation}
    d_i = (x^i_1, y^i_1, x^i_2, y^i_2, p_{obj}^i, p^i_1, …, p^i_C) 
\end{equation}
The prediction is encoded as a vector $d_i$ that consists of:
\begin{itemize}
    \item Bounding box information: $(x^i_1, y^i_1, x^i_2, y^i_2)$ denotes the top-left and bottom-right corners of the predicted box.
    \item Objectness probability score: $p_{obj}^i \in [0, 1]$ denotes the probability of an object's occurrence in the predicted box.
    \item Class score information: $(p^i_1, …, p^i_C)$ denotes the probability of $C$ classes in predicted box.
\end{itemize}

In almost all object detectors, such as Faster-RCNN~\cite{ren2015faster}, YOLOX~\cite{ge2021yolox}, the anchor boxes technique is widely used to detect bounding boxes.
G-CAME utilizes this technique to find and estimate the region related to the predicted box.
Our method can be divided into 4 phases (Fig.~\ref{fig:method_overview}) as follows: 1) Choosing target layers, 2) Object Locating, 3) Weighting Feature Map, and 4) Masking Target Region.

\subsection{Target layers selection}~\label{ss:target_layer}
\textbf{One-stage object detector (YOLOX)} For a one-stage object detector, such as YOLOX, we choose the final convolution layer in each branch of the model as the target layer to calculate the derivative, as convolutional layers naturally retain spatial information that is lost in fully connected layers. Hence, the last convolutional layers are expected to have the best compromise between high-level semantics and detailed spatial information~\cite{selvaraju2017grad}. The neurons in these layers look for semantic class-specific information in the image.

\textbf{Two-stage object detector (Faster-RCNN)} Two-stage object detectors, such as Faster-RCNN, contain two phases. 
In the first stage, the image is passed through stacked convolution layers in backbone layers and the Feature Pyramid Network (FPN)~\cite{lin2017feature}  which includes four branches to detect the different objects' sizes to extract features.
Subsequently, the Region Proposal Network (RPN) identifies potential object-containing regions, which are then resized uniformly via the Region of Interest (ROI) Pooling layer.
For a two-stage object detector, we utilize the convolution layers in the FPN network as the target layers to analyze because they are the last layers containing spatial information of the feature extractors.




\subsection{Object Localization with Gradient}
\label{subsec_1}
Most detector models like Faster-RCNN~\cite{ren2015faster}, PAFNet~\cite{xin2021pafnet} use the anchor box technique to predict the bounding boxes. However, regarding the YOLOX~\cite{ge2021yolox}, an anchor-free detector, in the final feature map, each pixel predicts $N$ bounding boxes 
and one bounding box for the anchor-free technique.
To get the correct pixel representing the box that we aim to explain, we take the derivative of the target box with the final feature map to get the location map $G_k^{l(c)}$ as the following formula:
\begin{equation} \label{gradient}
    G_k^{l(c)} = \frac{\partial S^c}{\partial A_k^l}
\end{equation}
where $G_k^{l(c)}$ denotes the gradient map of layer $l$ for feature map $k$.
$\frac{\partial S^c}{\partial A_k^l}$ is the derivative of the target class score $S^c$ with the feature map $A_k$.
In the regression task of most one-stage object detectors, $1\times1$ Convolution is used for predicting the bounding box, so in the backward pass, we have the Gradient map $G$ having the value of 1 pixel.

In the two-stage object detector, such as Faster-RCNN, because the regression and classification tasks are in two separate branches, we tailor G-CAME for two-stage models as follows.
First, we calculate the partial derivative of the class score according to each feature map of selected layers.
Faster-RCNN has four branches of detecting objects, and we choose the last convolution layer of each branch to calculate the derivative.
When we take the derivative of the class score to the target layer, the gradient map $G_k^{l(c)}$ has more than one pixel having value because anchor boxes are created in the next phase, namely the detecting phase.
The ROI pooling layer replaces 1$\times$1 Convolution, and they are in a separate branch from the classification stage.
Thus, we cannot get the pixel representing the object's center through the gradient map.
To solve this issue, we set the pixel with the highest value in the gradient map as the center of the Gaussian mask.
We estimate that the area around the highest value pixel likely contains relevant features.

\subsection{Weighting Feature Map via Gradient-based method}
\label{subsec_2}
We adopt a gradient-based method as GradCAM~\cite{selvaraju2017grad} for the classification to get the weight for each feature map. 
As the value in the gradient map can be either positive or negative, we divide all $k$ feature maps into two parts ($k_1$ and $k_2$, $k_1 + k_2 = k$), the one with positive gradient $A_k^{c(+)}$ and another with negative gradient $A_k^{c(-)}$. 
$\alpha_k^c$ is the weight for each feature map $k$ of target layer $l$ calculated by taking the mean value of the gradient map $G_k^{l(c)}$.
The negative $\alpha$ is considered to reduce the target score, so we sum two parts separately and then subtract the negative part from the positive one (as Eq.~\ref{our_CAM}) to get a smoother saliency map, and then use the $ReLU$ function to remove the pixel not contributing to the prediction.
\begin{equation} \label{neg_A}
    A_{k_2}^{c(-)} = \alpha_{k_2}^{c(-)}A_{k_2}^c
\end{equation}
\begin{equation} \label{pos_A}
    A_{k_1}^{c(+)} = \alpha_{k_1}^{c(+)}A_{k_1}^c
\end{equation}
\begin{equation} \label{our_CAM}
    L^c_{\text{CAM}} = ReLU \bigg (\sum_{k_1} A_{k_1}^{c(+)} - \sum_{k_2} A_{k_2}^{c(-)} \bigg )
\end{equation}

Because GradCAM can only explain classification models, it highlights all objects of the same class $c$.
By detecting the target object's location, we can tailor G-CAME to the object detection problem by explaining only one target object.

\subsection{Masking Target Region with Gaussian Distribution}
\label{subsec_3}
To deal with the localization issue, we propose to use  Gaussian distribution to estimate the region around the object's center.
Because the gradient map shows the target object's location, we estimate the object region around the pixel representing the object's center by using a Gaussian mask as the weight for each pixel in the weighted feature map $k$. 
The Gaussian kernel is defined as:
\begin{equation} \label{gauss}
    G_\sigma = \frac{1}{2\pi\sigma^2} \exp^{-\frac{(x^2 + y^2)}{2\sigma^2}}
\end{equation}
where the term $\sigma$ is the standard deviation of the value in the Gaussian kernel and controls the kernel size $\kappa$.
$x$ and $y$ are two linear-space vectors filled with value in range $[1, \kappa]$ one vertically and another horizontally. 
The bigger $\sigma$ is, the larger highlighted region we get. 
For each feature map $k$ in layer $l$, we apply the Gaussian kernel to get the region of the target object and then sum all these weighted feature maps. 
In general, we slightly adjusted the weighting feature map (Eq.~\ref{our_CAM}) to get the final saliency map as shown in Eq.~\ref{GCAM}:
\begin{equation}
\begin{aligned}
    L^c_{\text{GCAME}} = ReLU\bigg (\sum_{k_1} G_{\sigma(k_1)} \odot A_{k_1}^{c(+)} - \sum_{k_2} G_{\sigma(k_2)} \odot A_{k_2}^{c(-)} \bigg )
    \label{GCAM}
\end{aligned}
\end{equation}

\subsubsection{Choosing $\sigma$ for Gaussian mask}
The Gaussian masks are applied to all feature maps, with the kernel size being the size of each feature map, and the $\sigma$ is calculated as in Eq.~\ref{sigma}.

\begin{equation} \label{R}
    R = \log \left| \frac{1}{Z} \sum_i \sum_j G_k^{l(c)}\right|
\end{equation}
\begin{equation} \label{S}
    S = \sqrt{\frac{H \times W}{h \times w}}
\end{equation}
\begin{equation} \label{sigma}
    \sigma = R\log{S} \times \frac{3}{\lfloor{\frac{\sqrt{h \times w}-1}{2}}\rfloor}
\end{equation}
where the $\sigma$ is combined by two terms. 
In the first term, we calculate the expansion factor with $R$ representing the importance of location map $G_k^{l(c)}$ and $S$ is the scale between the original image size ($H\times W$) and the feature map size ($h \times w$). We use the logarithm function to adjust the value of the first term so that its value can match the size of the gradient map.
For multi-scale object detectors, we have a different $S$ for each scale level. 
In the second term, we choose Gaussian kernel size based on the $3\sigma$-rule~\cite{pukelsheim1994three} as the Eq.~\ref{org_gauss} and take the inverse value.

\begin{equation} \label{org_gauss}
    \kappa = 2 \times {\lceil{3\sigma}\rceil} + 1
\end{equation}

\subsubsection{Gaussian mask generation}
We generate each Gaussian mask with the following steps:
\begin{enumerate}
    \item Create a grid filled with value in range $[0, w]$ for the width and $[0, h]$ for the height ($w$ and $h$ is the size of the location map $G_k^{l(c)}$).
    \item Subtract the grid with value in position $(i_t, j_t)$ where $(i_t, j_t)$ is the center pixel of the target object on the location map.
    \item Apply Gaussian formula (Eq.~\ref{gauss}) with $\sigma$ as the expansion factor as Eq.~\ref{sigma} to get the Gaussian distribution for all values in the grid.
    \item Normalize all values in range $[0,1]$.
\end{enumerate}
By normalizing all values in range $[0,1]$, Gaussian masks only keep the region relating to the object we aim to explain and remove other unrelated regions in the weighted feature map.

\section{Experiments and Results}\label{sec:results}
We performed our experiment on the MS-COCO 2017~\cite{lin2014microsoft} dataset with 5000 validation images.
The models in our experiment are YOLOX-l (one-stage model) and Faster-RCNN (two-stage model). 
All experiments and conducted on NVIDIA Tesla P100 GPU.
G-CAME's inference time depends on the number of feature maps in selected layer $l$.
Our experiments run on model YOLOX-l with 256 feature maps for roughly 0.5 second per object.

\subsection{Sanity check}
\begin{figure} [tbh!]
    \centering
    \includegraphics[scale=0.5]{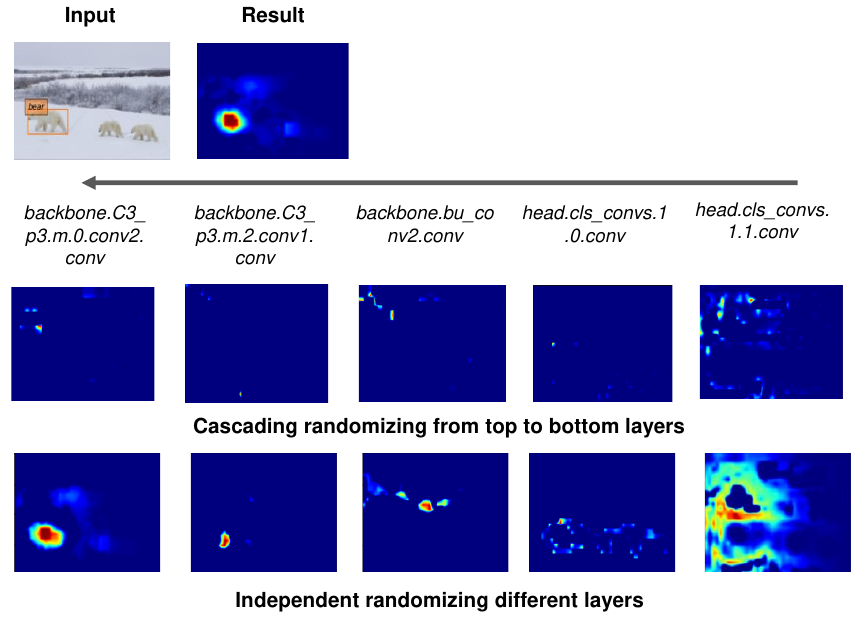}
    \caption{The result of Cascading Randomization and Independent Randomization for five layers from top to bottom of the YOLOX model. Chosen layers in the head part do not include the layer in the regression branch. The result shows G-CAME is sensitive to the model's parameters.}
    \label{fig:sanity_check}
\end{figure}

To validate whether the saliency map is a faithful explanation or not, we perform a sanity check~\cite{adebayo2018sanity} with Cascading Randomization and Independent Randomization. 
In the Cascading Randomization approach, we randomly choose five convolution layers as the test layers.
Then, for each layer between the selected layer and the top layer, we remove the pre-trained weights, reinitialize with normal distribution, and perform G-CAME to get the explanation for the target object.
In contrast to Independent Randomization, we only reinitialize the weight of the selected layer and retain other pre-trained weights.
The sanity check results show that G-CAME is sensitive to model parameters and can produce valid results, as shown in Fig.~\ref{fig:sanity_check}. 

\subsection{Qualitative Evaluation}
We performed a saliency map qualitative evaluation of G-CAME in comparison with D-RISE.
We use D-RISE's default parameters~\cite{petsiuk2021black}, where each grid's size is $16\times16$, the probability of each grid's occurrence is $0.5$, and the amount of samples for each image is $4000$.
For G-CAME, we choose the target layers as shown in Sec.~\ref{ss:target_layer} to calculate the derivative.

\begin{figure*}[tbh!]
    \centering
    \includegraphics[width=\linewidth]{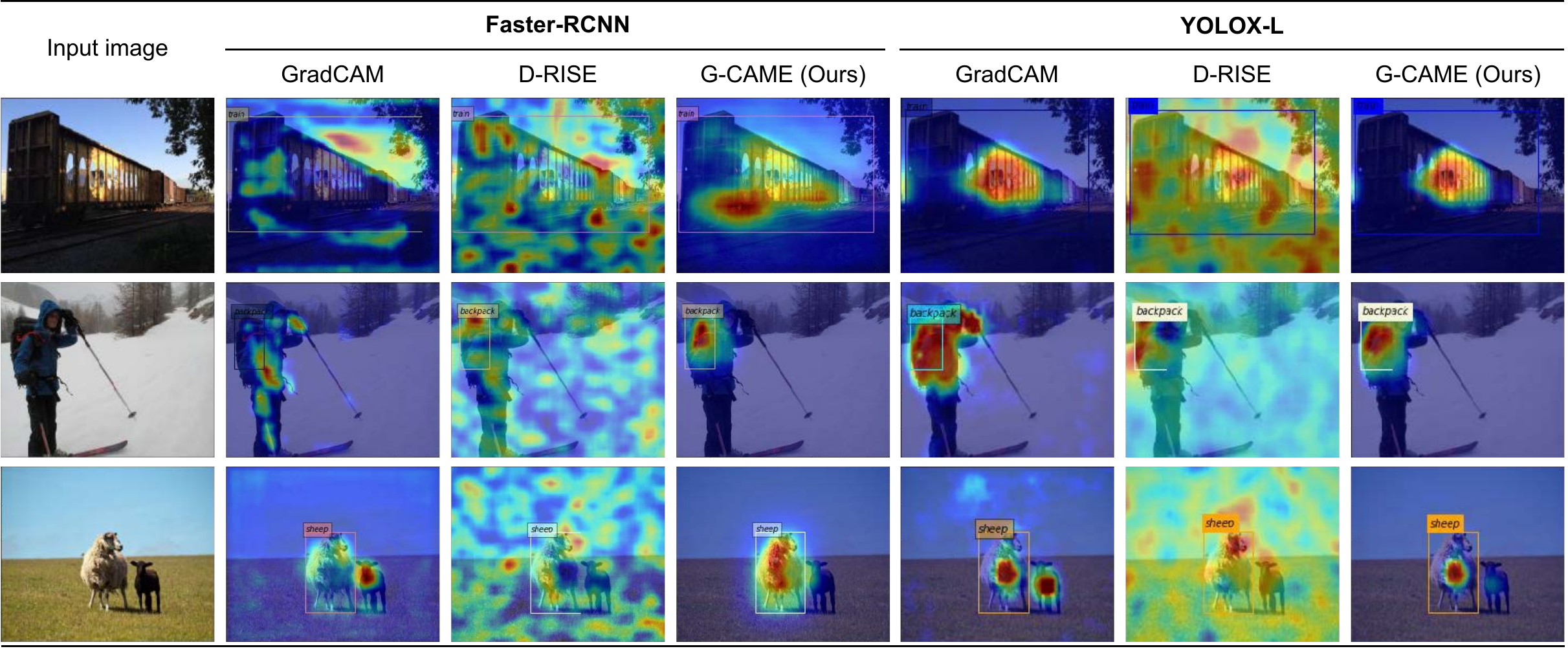}
    \caption{Visualization results of GradCAM, D-RISE, and G-CAME on samples of MS-COCO 2017 dataset. G-CAME can generate the least noisy saliency maps for explaining a specific object.}
    \label{fig:compare_rs}
\end{figure*}

Fig.~\ref{fig:compare_rs} shows the results of G-CAME compared with GradCAM and D-RISE.
GradCAM is only applicable for the classification task, as it shows the saliency maps for all objects in the same class.
Considering XAI methods for object detectors, where G-CAME and D-RISE can deliver the explanations for a specific object, G-CAME can generate saliency maps where the random noises are significantly reduced in comparison with D-RISE.


\subsection{Quantitative Localization Evaluation}
We use two standard metrics, Pointing Game \cite{zhang2018top} and Energy-based Pointing Game~\cite{wang2020score}, to compare the correlation between an object's saliency map and human-labeled ground truth. 
The results are shown in Table~\ref{tab:combined_rs}.

\subsubsection{Pointing Game (PG)}
To evaluate XAI methods via PG metric, firstly, we run the model on the dataset and get the bounding boxes that best match the ground truth for each class on each image.
A $hit$ is scored if the highest point of the saliency map lies inside the ground truth; otherwise, a $miss$ is counted. 
The pointing game score for each image is calculated by 
\begin{equation}
    PG = \frac{{\#}Hits}{{\#}Hits + {\#}Misses}
\end{equation}
This score should be high for a good explanation to evaluate an XAI method.

\subsubsection{Energy-Based Pointing Game (EBPG)}
EBPG~\cite{wang2020score} calculates how much the energy of the saliency map falls inside the bounding box.
Similar to the PG score, a good explanation is considered to have a higher EBPG.
EBPG formula is defined as follows:
\begin{equation} \label{ebpg}
    EBPG = \frac{\sum L^c_{(i, j) \in bbox}}{L^c_{(i, j) \in bbox} + L^c_{(i, j) \notin bbox}}
\end{equation}

PG and EBPG results are reported in Table~\ref{tab:combined_rs}.
Specifically, more than 65\% energy of G-CAME's saliency map falls into the ground truth bounding box compared with only 18.4\% of D-RISE.
In other words, G-CAME drastically reduces noises in the saliency map. 
In PG evaluation, G-CAME also gives better results than D-RISE. 98\% of the highest pixel lie inside the correct bounding box, while this number in D-RISE is 86\%.

\begin{table}[ht]
    \resizebox{.9\textwidth}{!}{%
    \begin{tabular}[t]{ccc}
        \toprule
        Method & D-RISE & G-CAME (Our) \\
        \midrule
        \begin{tabular}[c]{@{}c@{}}PG\%$\uparrow$\\(Overall \textbar{} Tiny object)\end{tabular} & 0.86 \textbar{} 0.127 & \textbf{0.98 \textbar{} 0.158} \\
        \midrule
        \begin{tabular}[c]{@{}c@{}}EBPG\%$\uparrow$\\(Overall \textbar{} Tiny object)\end{tabular} & 0.184 \textbar{} 0.009 & \textbf{0.671 \textbar{} 0.261} \\
        \bottomrule
    \end{tabular}
    \hspace{1pt}
    \begin{tabular}[t]{ccc}
    \toprule
        Method & D-RISE & G-CAME (Our) \\
        \midrule
        Confidence Drop\%$\uparrow$ & \textbf{42.3} & 36.8 \\
        \midrule
        Information Drop\%$\downarrow$ & 31.58 & \textbf{29.15} \\
        \midrule
        Running time(s)$\downarrow$ & 252 & \textbf{0.435} \\
        \bottomrule
    \end{tabular}
    }
    \caption{Comparison of D-RISE and G-CAME (Our) on the MS-COCO 2017 validation dataset with the YOLOX model. Evaluation metrics include PG\%, EBPG\%, Confidence Drop, Information Drop, and Running time. Higher or lower scores are better as indicated by $\uparrow/\downarrow$. The best results are shown in bold.}
    \label{tab:combined_rs}
\end{table}

\subsubsection{Bias in Tiny Object Detection}
\begin{figure}[h!]

    \centering
    \includegraphics[width=.8\linewidth]{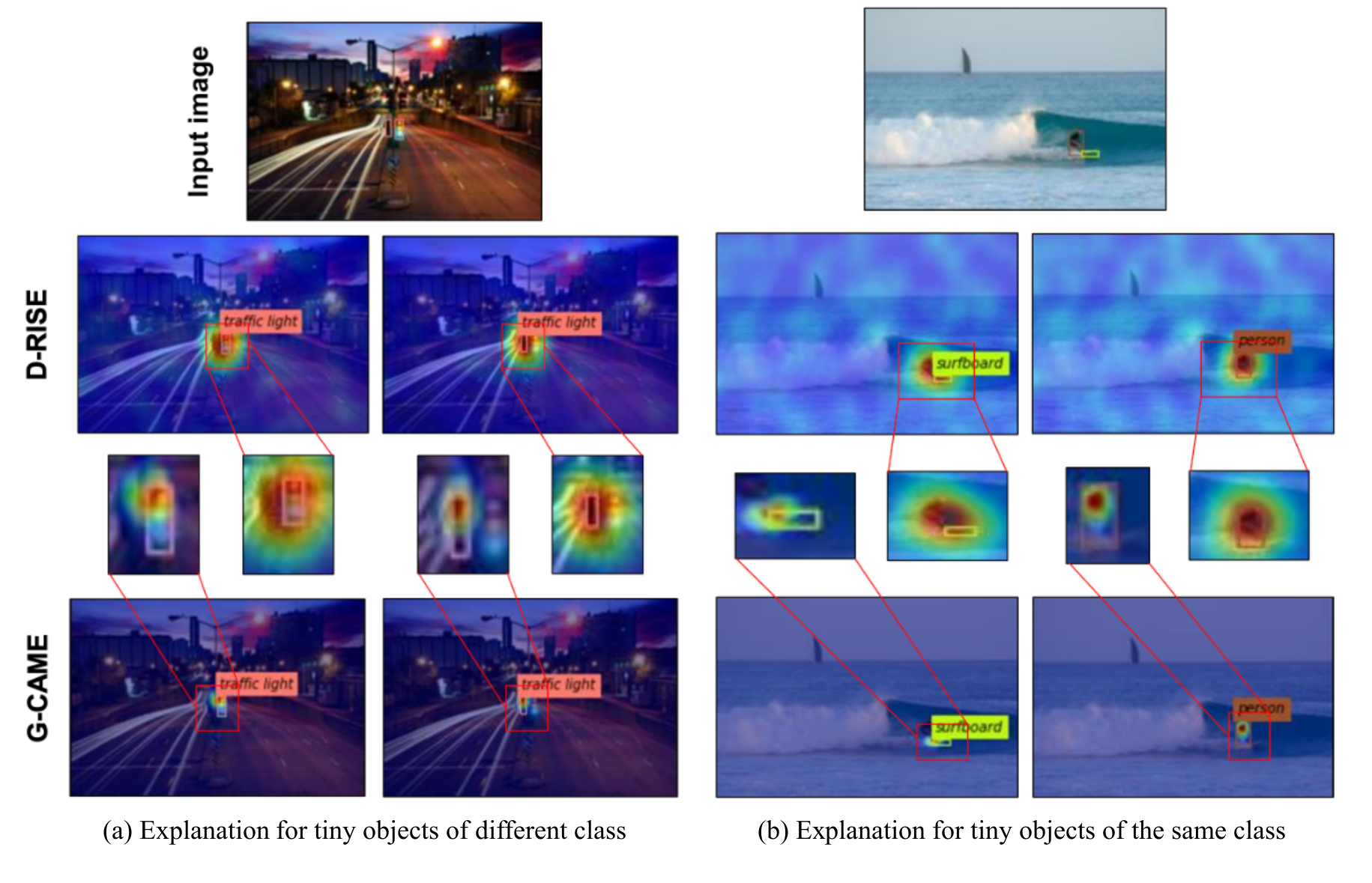}
    \caption{The saliency map of D-RISE and G-CAME for tiny objects prediction. We evaluate them in two cases: (a) multiple tiny objects from the same class lying close together and (b) multiple tiny objects from different classes lying close together. In both cases, G-CAME can clearly identify each object in its explanations.}
    \label{fig:tiny_obj}
\end{figure}

Explaining tiny objects detected by the model can be a challenge for XAI methods.
In particular, the saliency map may be biased toward the neighboring region. 
This issue can worsen when multiple tiny objects partially or fully overlap because the saliency map stays in the same location for every object. 
In our experiments, we define the tiny object by calculating the ratio of the predicted bounding box area to the input image area (640$\times$640 in YOLOX).
An object is considered tiny when this ratio is less than or equal to 0.005.
In Fig.~\ref{fig:tiny_obj}, we compare G-CAME with D-RISE in explaining tiny object prediction for two cases. 
In the first case (Fig.~\ref{fig:tiny_obj}a), we test the performance of D-RISE and G-CAME in explaining two tiny objects of the same class.
The result shows that D-RISE fails to distinguish two ``traffic lights'', where the saliency maps are nearly identical.
For the case of multiple objects with different classes overlapping (Fig.~\ref{fig:tiny_obj}b), the saliency maps produced by D-RISE hardly focus on one specific target.
The saliency corresponding to the ``surfboard'' even covers the ``person'', and so does the explanation of the ``person''.
The problem can be the grid's size in D-RISE, but changing to a much smaller grid's size can make the detector unable to predict. 
In contrast, G-CAME can clearly show the target object's localization in both cases and reduce the saliency map's bias to unrelated regions. 
In detail, we evaluated our method only in explaining tiny object prediction with EBPG score.
The MS-COCO 2017 validation dataset has more than 8000 tiny objects, and the results are reported in Table~\ref{tab:combined_rs}. 
Our method outperforms D-RISE with more than 26\% energy of the saliency map falling into the predicted box, while this figure in D-RISE is only 0.9\%. 
Especially, most of the energy in D-RISE's explanation does not focus on the correct target.
In the PG score, instead of evaluating one pixel, we assess all pixels having the same value as the pixel with the highest value.
The result also shows that G-CAME's explanation has better accuracy than D-RISE's.

\subsection{Quantitative Faithfulness Evaluation}
Another essential aspect of an XAI method is the ability to ensure the explanation's completeness and consistency in the model's predictions.
In this section, we employ the Confidence Drop and Information Drop scores to evaluate G-CAME and D-RISE on the YOLOX model with the MS-COCO 2017 dataset.

\subsubsection{Confidence Drop} 
We employ the Average Drop metric to evaluate the confidence change~\cite{chattopadhay2018grad, fu2020axiom, ramaswamy2020ablation} in the model's prediction for the target object when using the explanation as the input.
In other words, when we remove these important regions, the confidence score of the target box should be dropped.
The Average Drop is defined as:
\begin{equation} \label{Avg_drop}
    AD = \frac{1}{N}\sum_{i=1}^N \frac{max(P_c(I_i) - P_c(\tilde{I_i}), 0)}{P_c(I_i)} \times 100
\end{equation}
where:
\begin{equation} 
    \tilde{I_o} = I \odot (1 - M_o) + \mu M_o
    \label{perturbed}
\end{equation}
\begin{equation} 
    P_c(\tilde{I}) = IoU(L_i, L_j) \cdot p_{c(L_j)}
    \label{prob}
\end{equation}

Here, we tailor the original formula of Average Drop for the object detection model.
In Eq.~\ref{perturbed}, we create a new input image masked by the explanation $M$ of G-CAME.
$\mu$ is the mean value of the original image.
With the value of $M$, we only keep 20\% of the pixel with the most significant value in the original explanation and set the rest as 0.
Then, we can minimize the explanation's noise, and the saliency map can focus on the regions most influencing the prediction. 
 
In Eq.~\ref{prob}, to compute probability $P_c(\tilde{I})$, we first calculate the pair-wise $IoU$ of the box $L_j$ predicted on perturbed image $\tilde{I}$ with the box $L_i$ predicted on the original image and take the one with the highest value.
After that, we multiply the first term with the corresponding class score $p_{c(L_j)}$ of the box.
In calculating $P_c(I_i)$, the $IoU$ equals 1, so the value remains the original confidence score. 
Hence, if the explanation is faithful, the confidence drop should increase.
However, removing several pixels can penalize the method of producing the saliency map that has connected and coherent regions.
Specifically, pixels representing the object's edges are more meaningful than others in the middle~\cite{kapishnikov2019xrai}.
For example, pixels representing the dog's tail are easier to recognize than others lying on the dog's body.

\subsubsection{Information Drop}
In addition to the Confidence Drop score, we measure the faithfulness of the method via the Information Drop score. 
We compare the information level of the \textit{bokeh} image by blurring images with focused salient regions.
To measure the \textit{bokeh} image's information, we use WebP~\cite{Webp} format and calculate the Information Drop score by taking the proportion of the compressed size of the \textit{bokeh} image to the original image~\cite{kapishnikov2019xrai}.

\subsection{Evaluation}
Table~\ref{tab:combined_rs} highlights the strengths of G-CAME compared to D-RISE. D-RISE achieves a 42.3\% Confidence Drop by spreading its saliency map across the image, leading to a significant but less targeted reduction in confidence. Conversely, G-CAME maintains focus on the target object, resulting in a lower confidence drop that signifies a precise and relevant explanation.
Crucially, G-CAME outperforms D-RISE in Information Drop with 29.1\% versus 31.58\%, indicating superior preservation of the original image's content. Additionally, our method offers a significant speed advantage, delivering explanations in under a second, as opposed to D-RISE's four-minute runtime.
These results demonstrate G-CAME's efficiency in providing focused, relevant, and quick explanations for object detection models.

\section{Conclusion}\label{sec:conclusions}
In this paper, we proposed G-CAME, a novel CAM-based XAI method elevating the Gaussian kernel to explain one-stage and two-stage object detection models.
The experiment’s results show that our method can plausibly explain the model’s predictions and reduce the bias in tiny object detection.
Moreover, our method’s runtime is relatively short, overcoming the time constraint of existing region-based methods and reducing the noise in the saliency map.

\section*{Acknowledgment}
This work was partially supported by the NBIF Talent Recruitment Fund (TRF2003-001) and the UNB-FCS Startup Fund (22-23 START UP/ H CAO).


\printbibliography[heading=subbibintoc]

@article{nguyen2023towards,
  title={Towards Trust of Explainable AI in Thyroid Nodule Diagnosis},
  author={Nguyen, Truong Thanh Hung and Truong, Van Binh and Nguyen, Vo Thanh Khang and Cao, Quoc Hung and Nguyen, Quoc Khanh},
  journal={arXiv preprint arXiv:2303.04731},
  year={2023}
}

@article{truong2023towards,
  title={Towards Better Explanations for Object Detection},
  author={Truong, Van Binh and Nguyen, Truong Thanh Hung and Nguyen, Vo Thanh Khang and Nguyen, Quoc Khanh and Cao, Quoc Hung},
  journal={arXiv preprint arXiv:2306.02744},
  year={2023}
}

@inproceedings{nguyen2021evaluation,
  title={Evaluation of explainable artificial intelligence: Shap, lime, and cam},
  author={Nguyen, Hung Truong Thanh and Cao, Hung Quoc and Nguyen, Khang Vo Thanh and Pham, Nguyen Dinh Khoi},
  booktitle={Proceedings of the FPT AI Conference},
  pages={1--6},
  year={2021}
}

@inproceedings{selvaraju2017grad,
  title={Grad-cam: Visual explanations from deep networks via gradient-based localization},
  author={Selvaraju, Ramprasaath R and Cogswell, Michael and Das, Abhishek and Vedantam, Ramakrishna and Parikh, Devi and Batra, Dhruv},
  booktitle={Proceedings of the IEEE international conference on computer vision},
  pages={618--626},
  year={2017}
}

@inproceedings{zhou2016learning,
  title={Learning deep features for discriminative localization},
  author={Zhou, Bolei and Khosla, Aditya and Lapedriza, Agata and Oliva, Aude and Torralba, Antonio},
  booktitle={Proceedings of the IEEE conference on computer vision and pattern recognition},
  pages={2921--2929},
  year={2016}
}

@article{petsiuk2018rise,
  title={Rise: Randomized input sampling for explanation of black-box models},
  author={Petsiuk, Vitali and Das, Abir and Saenko, Kate},
  journal={arXiv preprint arXiv:1806.07421},
  year={2018}
}

@article{fu2020axiom,
  title={Axiom-based grad-cam: Towards accurate visualization and explanation of cnns},
  author={Fu, Ruigang and Hu, Qingyong and Dong, Xiaohu and Guo, Yulan and Gao, Yinghui and Li, Biao},
  journal={arXiv preprint arXiv:2008.02312},
  year={2020}
}

@inproceedings{petsiuk2021black,
  title={Black-box explanation of object detectors via saliency maps},
  author={Petsiuk, Vitali and Jain, Rajiv and Manjunatha, Varun and Morariu, Vlad I and Mehra, Ashutosh and Ordonez, Vicente and Saenko, Kate},
  booktitle={Proceedings of the\\ IEEE/CVF Conference on Computer Vision and Pattern Recognition},
  pages={11443--11452},
  year={2021}
}

@article{sejr2021surrogate,
  title={Surrogate Object Detection Explainer (SODEx) with YOLOv4 and LIME},
  author={Sejr, Jonas Herskind and Schneider-Kamp, Peter and Ayoub, Naeem},
  journal={Machine Learning and Knowledge Extraction},
  volume={3},
  number={3},
  pages={662--671},
  year={2021},
  publisher={MDPI}
}

@inproceedings{ribeiro2016should,
  title={" Why should i trust you?" Explaining the predictions of any classifier},
  author={Ribeiro, Marco Tulio and Singh, Sameer and Guestrin, Carlos},
  booktitle={Proceedings of the 22nd ACM SIGKDD international conference on knowledge discovery and data mining},
  pages={1135--1144},
  year={2016}
}

@inproceedings{chattopadhay2018grad,
  title={Grad-cam++: Generalized gradient-based visual explanations for deep convolutional networks},
  author={Chattopadhay, Aditya and Sarkar, Anirban and Howlader, Prantik and Balasubramanian, Vineeth N},
  booktitle={2018 IEEE winter conference on applications of computer vision (WACV)},
  pages={839--847},
  year={2018},
  organization={IEEE}
}

@inproceedings{redmon2016you,
  title={You only look once: Unified, real-time object detection},
  author={Redmon, Joseph and Divvala, Santosh and Girshick, Ross and Farhadi, Ali},
  booktitle={Proceedings of the IEEE conference on computer vision and pattern recognition},
  pages={779--788},
  year={2016}
}

@inproceedings{liu2016ssd,
  title={Ssd: Single shot multibox detector},
  author={Liu, Wei and Anguelov, Dragomir and Erhan, Dumitru and Szegedy, Christian and Reed, Scott and Fu, Cheng-Yang and Berg, Alexander C},
  booktitle={European conference on computer vision},
  pages={21--37},
  year={2016},
  organization={Springer}
}

@inproceedings{lin2017focal,
  title={Focal loss for dense object detection},
  author={Lin, Tsung-Yi and Goyal, Priya and Girshick, Ross and He, Kaiming and Doll{\'a}r, Piotr},
  booktitle={Proceedings of the IEEE international conference on computer vision},
  pages={2980--2988},
  year={2017}
}

@article{ren2015faster,
  title={Faster r-cnn: Towards real-time object detection with region proposal networks},
  author={Ren, Shaoqing and He, Kaiming and Girshick, Ross and Sun, Jian},
  journal={Advances in neural information processing systems},
  volume={28},
  year={2015}
}

@inproceedings{lin2017feature,
  title={Feature pyramid networks for object detection},
  author={Lin, Tsung-Yi and Doll{\'a}r, Piotr and Girshick, Ross and He, Kaiming and Hariharan, Bharath and Belongie, Serge},
  booktitle={Proceedings of the IEEE conference on computer vision and pattern recognition},
  pages={2117--2125},
  year={2017}
}

@article{dai2016r,
  title={R-fcn: Object detection via region-based fully convolutional networks},
  author={Dai, Jifeng and Li, Yi and He, Kaiming and Sun, Jian},
  journal={Advances in neural information processing systems},
  volume={29},
  year={2016}
}

@inproceedings{girshick2014rich,
  title={Rich feature hierarchies for accurate object detection and semantic segmentation},
  author={Girshick, Ross and Donahue, Jeff and Darrell, Trevor and Malik, Jitendra},
  booktitle={Proceedings of the IEEE conference on computer vision and pattern recognition},
  pages={580--587},
  year={2014}
}

@article{ge2021yolox,
  title={Yolox: Exceeding yolo series in 2021},
  author={Ge, Zheng and Liu, Songtao and Wang, Feng and Li, Zeming and Sun, Jian},
  journal={arXiv preprint arXiv:2107.08430},
  year={2021}
}

@article{zhang2018top,
  title={Top-down neural attention by excitation backprop},
  author={Zhang, Jianming and Bargal, Sarah Adel and Lin, Zhe and Brandt, Jonathan and Shen, Xiaohui and Sclaroff, Stan},
  journal={International Journal of Computer Vision},
  volume={126},
  number={10},
  pages={1084--1102},
  year={2018},
  publisher={Springer}
}

@inproceedings{wang2020score,
  title={Score-CAM: Score-weighted visual explanations for convolutional neural networks},
  author={Wang, Haofan and Wang, Zifan and Du, Mengnan and Yang, Fan and Zhang, Zijian and Ding, Sirui and Mardziel, Piotr and Hu, Xia},
  booktitle={Proceedings of the IEEE/CVF conference on computer vision and pattern recognition workshops},
  pages={24--25},
  year={2020}
}

@inproceedings{lin2014microsoft,
  title={Microsoft coco: Common objects in context},
  author={Lin, Tsung-Yi and Maire, Michael and Belongie, Serge and Hays, James and Perona, Pietro and Ramanan, Deva and Doll{\'a}r, Piotr and Zitnick, C Lawrence},
  booktitle={European conference on computer vision},
  pages={740--755},
  year={2014},
  organization={Springer}
}

@INBOOK{8689279,
  author={Grami, Ali},
  booktitle={Probability, Random Variables, Statistics, and Random Processes: Fundamentals \& Applications}, 
  title={The Gaussian Distribution}, 
  year={2019},
  volume={},
  number={},
  pages={201-238},
  doi={10.1002/9781119300847.ch7},
  ISSN={},
  publisher={Wiley},
  isbn={},
  url={https://ieeexplore.ieee.org/document/8689279},}

@inproceedings{ramaswamy2020ablation,
  title={Ablation-cam: Visual explanations for deep convolutional network via gradient-free localization},
  author={Ramaswamy, Harish Guruprasad and others},
  booktitle={Proceedings of the IEEE/CVF Winter Conference on Applications of Computer Vision},
  pages={983--991},
  year={2020}
}

@article{adebayo2018sanity,
  title={Sanity checks for saliency maps},
  author={Adebayo, Julius and Gilmer, Justin and Muelly, Michael and Goodfellow, Ian and Hardt, Moritz and Kim, Been},
  journal={Advances in neural information processing systems},
  volume={31},
  year={2018}
}

@article{xin2021pafnet,
  title={Pafnet: An efficient anchor-free object detector guidance},
  author={Xin, Ying and Wang, Guanzhong and Mao, Mingyuan and Feng, Yuan and Dang, Qingqing and Ma, Yanjun and Ding, Errui and Han, Shumin},
  journal={arXiv preprint arXiv:2104.13534},
  year={2021}
}

@inproceedings{kapishnikov2019xrai,
  title={Xrai: Better attributions through regions},
  author={Kapishnikov, Andrei and Bolukbasi, Tolga and Vi{\'e}gas, Fernanda and Terry, Michael},
  booktitle={Proceedings of the IEEE/CVF International Conference on Computer Vision},
  pages={4948--4957},
  year={2019}
}

@misc{Webp,
title = {Google, WebP format},
howpublished={\url{https://developers.google.com/speed/webp}}
}

@article{zhang2021group,
  title={Group-CAM: group score-weighted visual explanations for deep convolutional networks},
  author={Zhang, Qinglong and Rao, Lu and Yang, Yubin},
  journal={arXiv preprint arXiv:2103.13859},
  year={2021}
}

@article{pukelsheim1994three,
  title={The three sigma rule},
  author={Pukelsheim, Friedrich},
  journal={The American Statistician},
  volume={48},
  number={2},
  pages={88--91},
  year={1994},
  publisher={Taylor \& Francis}
}

\end{document}